\documentclass[sigconf]{acmart}

\usepackage{booktabs} 
\usepackage{todonotes} 

\setcopyright{rightsretained}
\usepackage{enumitem}
\usepackage{amsmath,amssymb,amsfonts}
\usepackage{algorithmic}
\usepackage{graphicx}
\usepackage{textcomp}
\usepackage{subcaption}

\def\BibTeX{{\rm B\kern-.05em{\sc i\kern-.025em b}\kern-.08em
    T\kern-.1667em\lower.7ex\hbox{E}\kern-.125emX}}
    
\copyrightyear{2018} 
\acmYear{2018} 
\setcopyright{acmlicensed}
\acmConference[FDG18]{Foundations of Digital Games 2018}{August 7--10, 2018}{Malm\"o, Sweden}
\acmBooktitle{Foundations of Digital Games 2018 (FDG18), August 7--10, 2018, Malm\"o, Sweden}
\acmPrice{15.00}
\acmDOI{10.1145/3235765.3235790}
\acmISBN{978-1-4503-6571-0/18/08}

\begin{document}

\title{AtDELFI:\\Automatically Designing Legible, Full Instructions For Games}
\author{Michael Cerny Green}
\email{mcgreentn@gmail.com}
\affiliation{%
  \institution{Tandon School of Engineering,\\ New York University}
  \city{New York City}
  \state{NY}
}
\author{Ahmed Khalifa}
\email{ahmed.khalifa@nyu.edu}
\affiliation{%
  \institution{Tandon School of Engineering,\\ New York University}
  \city{New York City}
  \state{NY}
}
\author{Gabriella A. B. Barros}
\email{gabbbarros@gmail.com}
\affiliation{%
  \institution{Tandon School of Engineering,\\ New York University}
  \city{New York City}
  \state{NY}
}

\author{Tiago Machado}
\email{tiago.machado@nyu.edu}
\affiliation{%
  \institution{Tandon School of Engineering,\\ New York University}
  \city{New York City}
  \state{NY}
}

\author{Andy Nealen}
\email{andy@nealen.net}
\affiliation{%
 \institution{Tandon School of Engineering,\\ New York University} 
 \city{New York City}
 \country{USA}
}

\author{Julian Togelius}
\email{julian@togelius.com}
\affiliation{%
  \institution{Tandon School of Engineering,\\ New York University}
  \city{New York City}
  \state{NY}
}

\ccsdesc[500]{Applied computing~Computer games}

\begin{abstract}
This paper introduces a fully automatic method for generating video game tutorials. The \emph{AtDELFI} system (AuTomatically DEsigning Legible, Full Instructions for games) was created to investigate procedural generation of instructions that teach players how to play video games. We present a representation of game rules and mechanics using a graph system as well as a tutorial generation method that uses said graph representation. We demonstrate the concept by testing it on games within the General Video Game Artificial Intelligence (GVG-AI) framework; the paper discusses  tutorials generated for eight different games. Our findings suggest that a graph representation scheme works well for simple arcade style games such as Space Invaders and Pacman, but it appears that tutorials for more complex games might require higher-level understanding of the game than just single mechanics.
\end{abstract}

\keywords{tutorial, video game, procedural content generation, artificial intelligence}

\maketitle
\renewcommand{\shortauthors}{M. Green et al}

\section{Introduction}

Artificial intelligence has been used in many roles in games, most prominently for playing games, generating game content, and modeling players~\cite{yannakakis2018artificial}. In this paper, we present a prototype system for generating video game tutorials, a new application of AI in games presenting both important potential practical applications and hard research challenges.

Tutorials are designed to help you learn to play a game. They come in several different forms, such as textual instructions (e.g. ``press A to jump''), videos where an agent plays part of the game to show how it's done, and instructional content such as levels that gradually introduce core mechanics. Many video games utilize some combination of these different types of tutorials to teach players how to play them.


Being able to automatically or semi-automatically generate game tutorials would have significant benefits for game development, given the cost and effort associated with authoring tutorials for games. But it would also be important for making the vision of automated game generation possible, as attempts at video game generation so far have highlighted the difficulty of evaluating generated games without knowing how to play them as a human~\cite{nielsen2015towards,cook2014ludus}. Finally, a system for automatic tutorial generation may also give us insight into game design itself, as it could show us new ways of playing a game or give us a way of measuring qualities of a game (such as its depth) from the generated tutorial~\cite{lantz2017depth}.

The system we describe in this paper generates a combination of textual instructions and visual videos. Given a simple arcade game, it describes a set of actions to take in order to score points and to win the game. The actions are at the level of a single game mechanic, such as firing a shot at something or collecting an item. Each instruction is complemented by a short video snippet, showing how a game-playing agent executes the action. 

In order to make this possible, the system described in this paper builds on top of the General Video Game AI Framework (GVG-AI)~\cite{perez20162014,perez2018general}, which is an AI benchmarking and game prototyping framework where games are specified in the Video Game Description Language (VGDL)~\cite{ebner2013towards}. VGDL is a high-level language for 2D arcade games, and using this language allows us to analyze games at the mechanics level. The GVG-AI framework also contains a number of good general-purpose game-playing agents, allowing us automatically play the games involved, another crucial part of the tutorial generation process. Finally, as the GVG-AI framework currently contains more than 100 games, we can easily test the system over a large number of different games.

While building our tutorial generation system on top of the GVG-AI framework can be seen as a form of ``cheating'', given the capabilities GVG-AI affords---in particular, access to the game rules in easy-to-parse symbolic format---and the fact that no released games are written in VGDL, we see it as a form of scaffolding that allows us to prototype a system to showcase the capabilities a truly general tutorial generation could one day include, once the requisite underlying technologies exist. We foresee that, in the future, some combination of automatic game state identification, code inspection and other technologies could make this system a reality.

\section{Background}


This section expands on the evolution of tutorials in video games and different types of tutorials. Additionally, it highlights works related to automatic methods for tutorial generation and finding meaning from game mechanics.

\subsection{An Overview of Tutorial Evolution}

In video games, tutorials are often the very first thing a player encounters. Their purpose is to teach a player how to play the game. 
Tutorials and how they are presented to players have been evolving since the first games were introduced to the public~\cite{therrien2011get}. 

The first electronic games often lacked formal tutorials. Arcade games had simple mechanics that players could pick up easily: ``Tilt the joystick to move,'' ``Press button to shoot,'' etc. \emph{Pong} (Atari 1972) is an example of a simplistic game without need of a tutorial. As arcade games became more popular, game designers began to include easy-to-read mechanic tables and videos on the arcade machines themselves. \emph{StreetFighter} (Capcom 1987) and \emph{Pac-Man} (Namco 1980) used both images and text written on the machine surface, as well as in-game videos of certain mechanics and moves players can use.

The introduction of consoles to the gaming community allowed designers to create games with more complex mechanics. To help players achieve their goals in the game, designers began to include formal tutorials teaching players the gameplay basics. Games like \emph{Space Invaders} (ATARI 1978) and \emph{Super Mario Bros} (Nintendo 1985)
included booklets with text and photos explaining certain moves and mechanics.

\begin{figure}
\centering
\begin{center}
	\includegraphics[width=\columnwidth]{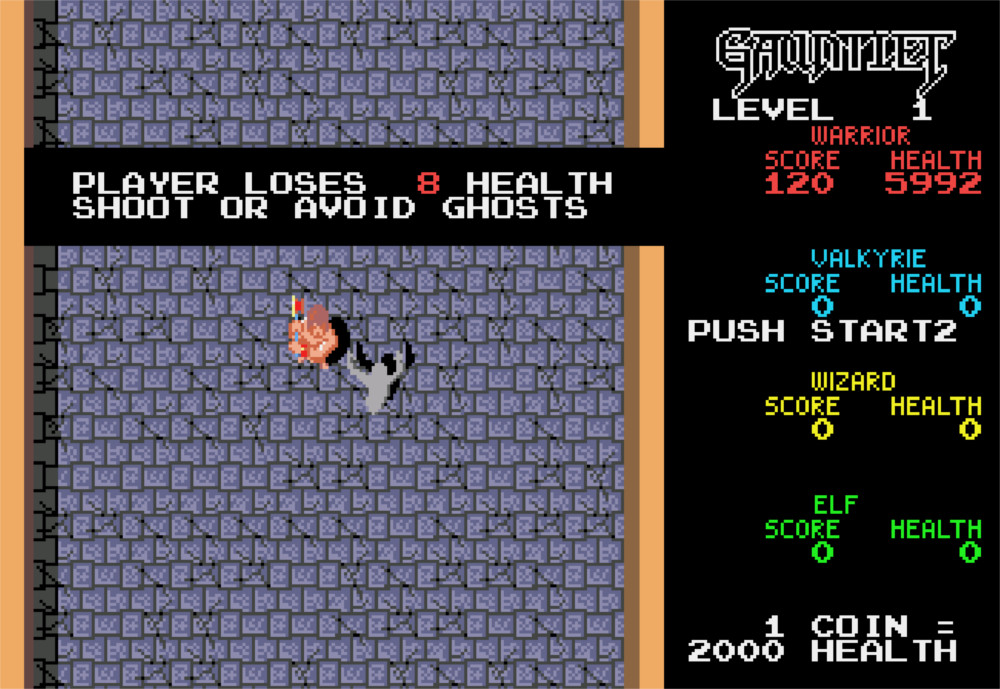}
\end{center}
\caption{Gauntlet interactive tutorial about colliding with enemies.}
\label{fig:gauntlet-tutorial}
\end{figure}

Designers experimented with other methods of teaching game rules, such as interactive in-game tutorials. Figure \ref{fig:gauntlet-tutorial} shows an early example of interactive in-game tutorial in \emph{Gauntlet} (Atari 1985), a complicated RPG dungeon crawler. To ease the player's learning curve, the tutorial was split among several levels. Whenever players first encountered a new element, the game paused and explanatory text appeared. Designers began to take advantage of just-in-time information systems that video games afforded, which has only in recent years been explored and researched in depth along with other similar techniques~\cite{gee2003video,plass2015foundations,gee2006video,moreno2007interactive}. 
Parametrized design spaces \cite{powley2017wevva,isaksen2017exploring} might assist with the exploration of interactive tutorial design, as it becomes easier to experiment with different teachable elements in a level.

\subsection{Tutorial Types and Content}

Whether or not a tutorial is ingrained in gameplay, it tends to fall somewhere along the axis of Sheri Graner Ray's knowledge acquisition styles for players~\cite{ray2010learning}. These styles range from Explorative Acquisition to Modeling Acquisition. \emph{Explorative acquisition} emphasizes learning about something \textit{by} doing it, whereas \emph{modeling acquisition} focuses on studying how to do something \textit{before} doing it. While one cannot say with absolute certainty if one technique is superior to another, different techniques suit different audiences and/or games~\cite{andersen2012impact,williams2009pedagogy,ray2010learning}. According to Andersen et al., the effectiveness of tutorials on gameplay depends on how complex a game is to begin with~\cite{andersen2012impact}; sometimes tutorials are not useful at all.

In a previous research, we offered a non-exhaustive list of three different tutorial types in games: Teaching Using Instruction, Teaching Using videos, and Teaching Using a Well Designed Experience~\cite{green2017press}. \emph{Instruction Tutorials} are simple text-based instructions that the player reads, similar to how board game instruction books explain rules. \emph{Demonstration Tutorials} show examples of actions the player can take in game, such as in \emph{Mega Man X} (Capcom 1993) when the player is shown the charging up mechanic~\cite{egoraptor2011megaman}. \emph{Carefully Designed Experience Tutorials} set up scenarios within which the player can explore and discover rules and mechanics. One of the most well known examples of this would be \emph{Super Mario Bros} World 1-1, a level that teaches the player about the jumping mechanic, that the goal of the game is to go to the right, what is good to collect in the game (coins and power ups), and what is bad (goombas).

\begin{figure}[tb]
\centering
\begin{center}
\includegraphics[height=0.2\textheight]{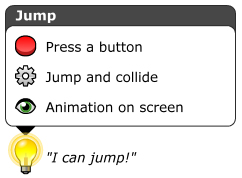}
\caption[Caption for LOF]{A skill atom for learning how to jump in any generic game, in the order of \textit{action} (button), \textit{simulation} (jump and collide), \textit{feedback} (animation on screen), and \textit{modeling} (``I can jump!'')\footnotemark}
\label{fig:skill-atom}
\end{center}
\end{figure}

To gain more understanding about tutorials, games, and mechanics, researchers have created various languages to model games. GVGAI's VGDL (discussed in \ref{sec:GVGAI}) is an example of such a language.
Dan Cook's concept of \emph{skill atoms}~\cite{cook2007chemistry}, which refers to the feedback loop through which a player learns a new skill during gameplay, is another example. Figure \ref{fig:skill-atom} shows the skill atom for learning how to jump. A skill atom can be divided into four separate elements:
\begin{itemize}
\item The \emph{Action} the player performs to learn a new skill. This could involve anything from pressing a button or doing a complex series of actions to accomplish an end goal.
\item The \emph{Simulation} of that action in game. The player's action somehow affects the world.
\item The \emph{Feedback} from the simulation informs the player of the new state of the game.
\item The \emph{Modeling} the player now performs within their head, mapping the action they just took to the feedback from the simulation. ``If I press this button, my character jumps up.''
\end{itemize}
Skill atoms can be associated with other skill atoms to form \emph{skill chains}. Using skill chains, one could presumably model any game.

A related concept is strategy ladders, where each step is an addition to the previous step's strategy that makes a significant difference in playing. It has been proposed that a game's depth can be defined as the length of its longest strategy ladder~\cite{lantz2017depth}. 

\subsection{Generative Methods for Tutorials and Game Mechanics}

Several systems address challenges related to tutorial generation. \emph{TutorialPlan}~\cite{li2013tutorialplan} generates text and image instructions for new users of AutoCAD. Blackjack and poker heuristics have been generated to create effective strategies for beginners\cite{de2016generating,de2018texas,de2018flop}. Mikami et al. generated tutorials for API coding libraries~\cite{mikami2014automatic}, claiming that the resulting generated tutorials helped users learn libraries more effectively than current tutorials. Alexander et al. formalized the game logic of \emph{Minecraft} (Mojang 2009) into mechanics to create action graphs, representing the player experience, and created quests and achievements based off those actions~\cite{alexander2017deriving}.

Another approach is that of Game-O-Matic~\cite{treanor2012game}, which generates arcade style games and instructions using a story-based concept-map inputted by a user. After the game is created, Game-O-Matic generates a tutorial page, explaining who the player will control, how to control them, and winning/losing conditions, by using the concept-map and relationships between objects within it.
\footnotetext{image from \url{https://www.gamasutra.com/view/feature/129948/the_chemistry_of_game_design.php?page=3}}
Mappy is a system which takes a Nintendo Entertainment System game and a sequence of buttons presses as input to generate an approximation of a linked map of rooms~\cite{osborn2017automatic}. Mappy essentially attempts to create understanding of map levels from movement mechanics. This is similar to what Summerville et al. created as a part of the \emph{Gemini} system, a logic program that performs static reasoning over game specifications in order to find meaning~\cite{summerville2017mechanics}. One can see overlap between our system and theirs, in particular the similarities between AtDelfi's condition/action nodes (covered in section \ref{sec:overview-nodes}) and Cygnus' precondition/result formalisms. Within the Cygnus system, the player can derive higher-level meanings about the game in question from implicit rules (which they call ``dynamics''). However this derived knowledge is not structured in the form of a tutorial. Whereas the Cygnus system takes game mechanics as input, AtDelfi was designed to perform automatic mechanic identification and automatically construct tutorials out of that.


\section{The General Video Game Artificial Intelligence framework}\label{sec:GVGAI}
The General Video Game Artificial Intelligence framework~\cite{perez20162014} (GVG-AI) is a framework built to run games written in the Video Game Description Language (VGDL)~\cite{ebner2013towards}. GVG-AI was originally developed in the context of the eponymous competition, but has since been used in various research projects
. 
The framework allows the use of automated agents interchangeably between games. To be well-equipped to play different games, an agent must hold a varying set of skills, such as reacting to the system, agile decision making and long-term planning. Thus, success at the set of GVG-AI games involves adapting to changing mechanics, goals and strategy requirements in parallel fashion to how high-skilled players adapt to their opponent strategies to win.

The Video Game Description Language~\cite{ebner2013towards} (VGDL) is a game description language used to represent $2$D games of the arcade (Pac-man), action (Space Invaders), and/or puzzle game (Sokoban) genres. VGDL games consist of $2$ parts: a game description and level descriptions. The game description is responsible for holding information about game objects, game mechanics and termination conditions.  Figure~\ref{fig:vgdl} shows a simple Sokoban game and its game description. The game description consists of $4$ subparts:

\begin{figure}[tb]
\centering
\begin{subfigure}[b]{0.49\textwidth}
\begin{center}
	\includegraphics[width=0.80\textwidth]{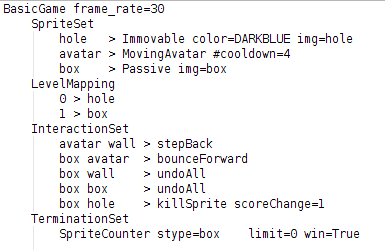}
    \end{center}
    \end{subfigure}
    \begin{subfigure}[b]{0.49\textwidth}
    \includegraphics[width=\textwidth]{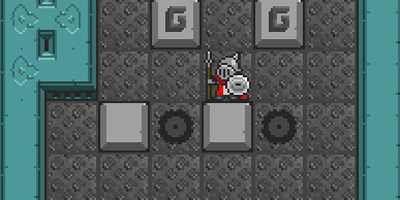}
    \end{subfigure}
     \caption{A definition of a simple game (a version of Sokoban) in VGDL, and a screenshot of part of the game in-engine.}
     \label{fig:vgdl}
\end{figure}

\begin{description}
\item [Sprite Set:] a hierarchical list of game objects, called game sprites. Similar game sprites can be grouped under a common name in the hierarchy, which is considered a parent of these similar game sprites. For example, in Pac-man, two sprites are grouped into the ``Pac-man'' parent type: \textit{hungry} and \textit{powered}. Each type shares some game rules with the other while also having different game rules associated with it: Whereas \textit{hungry} can be destroyed by ghosts, \textit{powered} can destroy ghosts instead. Each sprite has a type, orientation, and image. The sprite type defines its behavior, e.g. in Pac-man, a ghost is a \emph{RandomPathAltChaser}, which means it chases a ``hungry'' Pac-man but flees from it after Pac-man eats a power pellet.
\item [Interaction Set:] a list of all game interactions. Interactions occur upon collision between two sprites. For example, In Pac-man, if the player collides with a pellet, the latter will be destroyed and the score increases.
\item [Termination Set:] a list of conditions, which define how to win or lose the game. These conditions can be dependent on sprites or on a countdown timer. For example, in Pac-man, if the player eats all the pellets, the player wins the game.
\item [Level Mapping:] a table of characters and sprite names that is used to decode the level description.
\end{description}

The level description contains a $2$D matrix of characters, and each character can be decoded using the Level Mapping. Each character maps to game sprites' starting location for that level. 



\section{System Overview}
\emph{AtDELFI}'s tutorial generation process starts by creating a ``mechanic graph'', which tracks all mechanics, game objects, winning/losing conditions, and user-input information in the game. It encapsulates objects, conditions, and actions as nodes and creates edges, which represent relationships between nodes, as described in Sections \ref{sec:overview-nodes} and \ref{sec:overview-points}. After building the graph, the system finds ``critical paths'' by tracing mechanics from user inputs all the way to terminal states, as explained in Section \ref{sec:overview-paths}. Occasionally, rules are similar enough to undergo ``rule merging'', which we explain in Section \ref{sec:overview-merge}. The system then constructs the tutorial instructions (see Section \ref{sec:overview-instructions}) and finally collects frames with agent-play-throughs to display on a \emph{tutorial card} (see Section \ref{sec:overview-videos}).

\subsection{The Mechanic Graph}\label{sec:overview-graph}

At the core of tutorial generation is the system's understanding of the game rules and mechanics. To organize this, the system uses a directed graph to keep track of relationships between elements of gameplay, which is inspired by Dormans' mission graph concept~\cite{dormans2010adventures,dormans2011level,dormans2011generating}. By representing the game space as a mechanic graph, one can easily trace mechanics and find skill chains described by Cook~\cite{cook2007chemistry}.
\subsubsection{Nodes}\label{sec:overview-nodes}

There are three unique types of nodes: \emph{objects}, \emph{conditions}, and \emph{actions}. Objects are anything the player can interact with or see in a game, e.g. characters, enemies and the player avatar. Time is also included as an in-game entity, as some games have termination conditions based on a certain amount of time passing. Actions are verbs the game uses to change in-game objects, such as destroying a specified game object, transforming a object into another object, or winning the game. Conditions mark events that must take place to enact an action, as, for example, pressing a button, checking if the amount of passed game time is over a certain amount, or if the player's sprite has been destroyed. 

\begin{figure}
  \begin{center}
      \includegraphics[height=0.13\textwidth]{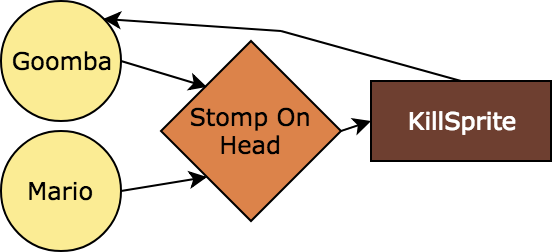}
  \end{center}
\caption{An example mechanic: ``If the Player stomps on the head of a Goomba, destroy the Goomba.''}
\label{fig:concept-mechanic}
\end{figure}

A set of nodes forms a mechanic, which describes the condition an object or group of objects must trigger for an associated action to take place. This representation is similar to Cook's skill atoms~\cite{cook2007chemistry}. When mechanics are linked together, they create structures like Cook's skill chain theory. Figure \ref{fig:concept-mechanic} is an example of a mechanic describing how stomping on the head of a Goomba in Super Mario Bros (Nintendo 1985) will destroy said Goomba. A ``Player'' object node and a ``Goomba'' object node both point at a ``Stomp On Head'' condition node. The condition node leads to a ``Destroy'' action node, which is pointing back to the ``Goomba'' object node.

\subsubsection{Controls and Points}\label{sec:overview-points}

Depending on the type of game being played, information about control and input may be contained within a node representing the player character, or stored separately from the graph in case the player is not explicitly controlling a single character. This type of information includes anything that explains the player's ability to affect the game through direct input, such as pressing buttons or using a game-pad.

Information about points is contained in the specific action that would cause an increase or decrease of points. For example, stomping on a Goomba in \emph{Super Mario Bros} (Nintendo 1985) gives the player 100 points, and would be represented by a mechanic similar to the one in Section \ref{sec:overview-nodes}. The only difference would be found in the ``Destroy'' action node, where there would be an additional attribute about raising the score by 100 points.

\subsubsection{Critical Paths}\label{sec:overview-paths}

Just as each node in the mechanic graph has inputs and outputs, mechanics themselves can be represented as having inputs and outputs. We define critical paths as the series of mechanics starting from an explicit player input and ending with a winning or losing terminal state. For example, a player wins \emph{Space Invaders} (Nishikado 1978) when all aliens in a level are destroyed. A critical path for \emph{Space Invaders} would be something like: ``Use the joystick to move, press A to shoot. If a missile and an alien collide, destroy the alien (and get points). If all aliens are destroyed, the player wins.'' The specific pathfinding algorithm used to find a correct critical path depends on the game and game mechanics. 

\subsubsection{Rule Merging}\label{sec:overview-merge}

Similar mechanics have the potential to be combined to simplify how tutorial instructions are presented to the user. We call this process ``rule merging.'' The exact implementation behind this depends almost entirely on the game, but the motivation is the same: make the tutorial more legible by grouping similar mechanics. For example, using Super Mario Bros as an example again, instead of saying ``the Player can kill Goombas using fireballs and the Player can kill Koopas using fireballs,'' the system can simplify it to ``the player can kill enemies using fireballs.''

\subsection{Generating Instructions}\label{sec:overview-instructions}

Using winning and losing critical paths, points, and movement information, the system uses a grammar to generate text instructions. Like rule merging, this process is heavily dependent on a given game, as language will vary based on particular game mechanics. For example, colliding with anything in \emph{Space Invaders} will require instruction language such as ``Colliding with an alien will destroy the alien.'' An item in \emph{The Legend of Zelda} (Nintendo 1986), on the other hand, will be ``collected'' if the player collides with it.

\subsection{Creating videos}\label{sec:overview-videos}

In order to gather video frames to display in a tutorial, the system captures play-through data from artificial agents who can reliably beat the game. By tracking when mechanics are triggered in-game, the system can then display an agent executing that mechanic next to the instructions describing it.

\section{Creating Tutorials for GVG-AI Games}

As a proof-of-concept for the \emph{AtDELFI} system, we tested our design on games from the GVG-AI framework. The following subsections describe the process of generating a tutorial for any GVG-AI game and include GVGAI's Aliens as an example of this process.

\subsection{Building the Graph from VGDL}\label{sec:creating-VGDL}

First, the system reads the Sprite Set, the Interaction Set, and the Termination Set from the game description file of a GVG-AI game. Every sprite found in the Sprite Set is given an \emph{object node} within the graph. Every interaction and termination in the Interaction and Termination Set is given a \emph{condition node} and an \emph{action node}.


In GVG-AI, it is safe to assume that any given interaction within the Interaction Set in a game will have a ``collide'' condition. However, the Termination Set might contain mechanics with unique conditions, such as counting how many sprites exist in the game or ending the game after a certain amount of time. Sprite nodes involved in an interaction or termination rule are linked in the directed graph to their associated condition nodes, while condition nodes are linked to their associated action nodes. This process' end result is a fully functional directed graph that adequately maps every mechanic in the game. A generated graph of GVGAI's Aliens has been included in the supplemental documents for this paper.
\subsubsection{Controls}

Information about user controls is implicitly stored when the avatar sprite (the sprite whom the player always has direct control of in any GVG-AI game) is registered into a sprite node, as all GVG-AI games require the player to explicitly control a single sprite. The avatar type determines controls and movement afforded to the player. Table \ref{table:avatar-movement} shows the different types of avatars and how movement information is parsed from them.
\begin{table}[tb]
\centering
\caption{Avatar Type and Movement Parsing}
\label{table:avatar-movement}
\begin{tabular}{l|l}
\textbf{Avatar Type }    & \textbf{Movement}                                                                                                                                  \\ \hline \hline
Moving          & ``...use the four arrow keys to move.''                                                                                                     \\ \hline
Horizontal      & ``...use the left and right arrow keys to move.''                                                                                           \\ \hline
Flak            & \begin{tabular}[c]{@{}l@{}}``...use the left and right arrow keys to move. \\ Press space to shoot/use/release {[}object{]}.''\end{tabular} \\ \hline
Vertical        & ``...use the up and down arrow keys to move.''                                                                                              \\ \hline
Ongoing         & \begin{tabular}[c]{@{}l@{}}``...use the arrow keys to change direction. \\ You will not stop traveling in that \\direction until you change direction again.''\end{tabular}                                                                                            \\ \hline
Ongoing Shoot   & \begin{tabular}[c]{@{}l@{}}``...use the arrow keys to change direction. \\ Press space to shoot/use/release {[}object{]}.''\end{tabular}    \\ \hline
Ongoing Turning & \begin{tabular}[c]{@{}l@{}}``...use the arrow keys to change direction. \\ You cannot do 180 degree turns!''\end{tabular}                   \\ \hline
Oriented        & ``...use the arrow keys to turn and move.''                                                                                                 \\ \hline
Shoot           & \begin{tabular}[c]{@{}l@{}}``...use the arrow keys to turn and move. \\ Press space to shoot/use/release {[}object{]}.''\end{tabular}       \\
\end{tabular}
\end{table}

\subsubsection{Points}

Every mechanic's action node contains score data. Each interaction in the Interaction Set that results in a change in points registered its mechanic as a point-changing mechanic in the graph's generation and stored how it affects point totals. Because of this, the graph has a list of all point-affecting mechanics.

\subsubsection{Critical Paths}

Critical paths can lead to wins or losses. For critical paths that lead to wins, the system observes every avatar sprite node in the graph (in some GVG-AI games, there are multiple avatar sprites). From each avatar sprite node, it then finds every possible path from that node to every winning terminal mechanic. 
Afterwards, it sorts these paths by length and picks the shortest path for every terminal mechanic. Finally, it iterates over every avatar sprite and picks the longest ``shortest path'' for each terminal mechanic. Preliminary testing suggested that this method results in a concise critical path that does not cut out important mechanics.

For critical paths leadings to losses, the system observes every losing terminal mechanic. From each of these mechanics, it works backwards attempting to decipher what might cause the mechanic to activate. In GVG-AI, there are two terminal conditions: a timeout, or the current number of clones of a certain sprite is at some threshold. The first is easy for the system to decipher, as no specific mechanic causes a timeout. The second requires the system to look at the sprite involved in the terminal mechanic, and see what other mechanics would impact its value. Any mechanic that achieves this goal is included in the losing critical path, i.e. any mechanic causing the number of clones of this sprite to rise or fall.


\subsection{Rule Merging}\label{sec:creating:rules}

\begin{figure}
\begin{center}
\includegraphics[width=\columnwidth]{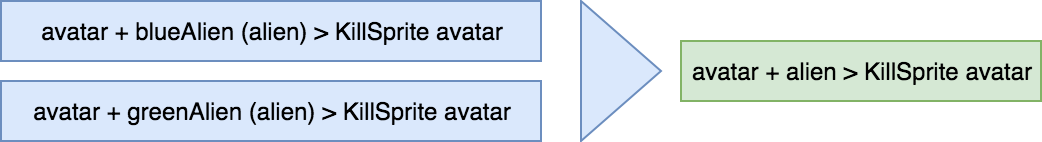}
\end{center}
\caption{Aliens - Avatar KillSprite Rule Merging}
\label{fig:rule-merging-aliens}
\end{figure}

There are cases where rules in a GVG-AI game are similar enough to be \emph{merged} when describing them to the player, i.e. combined into a single rule rather than explained separately. For every rule from the points and critical paths sections, the system loops through sprites that share parents with the two sprites involved in the mechanic. If it can find an identical mechanic in every sprite, the mechanics are merged into a single mechanic. Parent data for a sprite can be found within the sprite node, as described in Section \ref{sec:creating-VGDL}. 
Figure~\ref{fig:rule-merging-aliens} provides an example of rule merging in the GVG-AI's Aliens. It shows two different rules being merged into one generic rule, which destroy the avatar upon collision with both kinds of alien. Since both alienGreen and alienBlue destroy the avatar upon collision, they can merge into a single ``alien''-based rule.

\subsection{Tutorial Instructions}\label{sec:creating-instruction}
\begin{figure}
\includegraphics[height=0.4\textwidth]{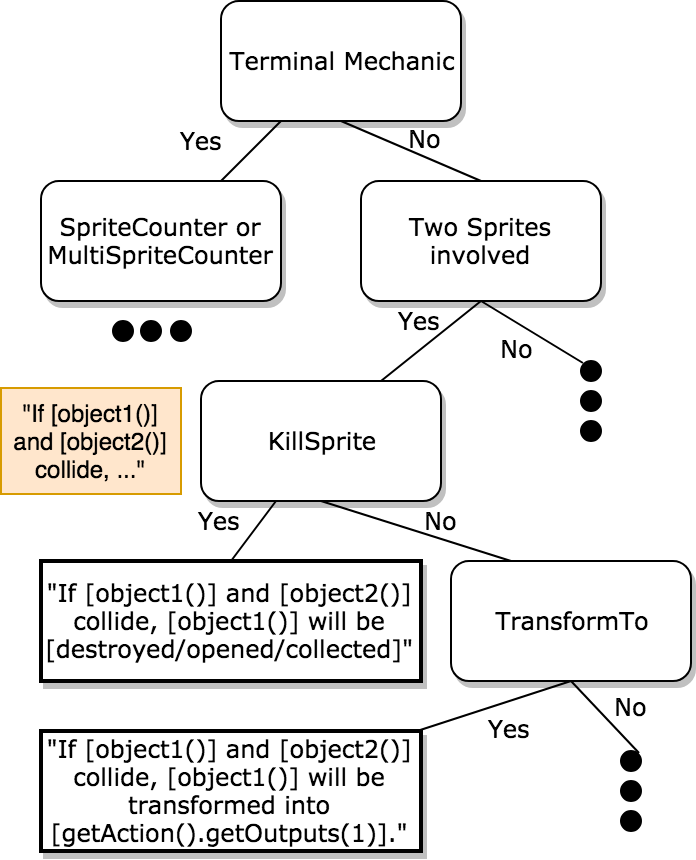}
\centering
\caption{A subtree of the overall text-replacement tree for all GVGAI games in the AtDELFI System}
\label{fig:text-replacement-tree}
\end{figure}

A tutorial's instructions are game explanations written in plain text. To create instructions, the system performs text-replacement using a decision tree. A subsection of this tree is displayed in Figure \ref{fig:text-replacement-tree}.
Using information about controls, points, and winning/losing critical paths, the system can generate human-readable text, which is displayed to the user. Table \ref{table:aliens-instructions} shows an example of generated tutorial instructions for GVG-AI's Aliens. It is important to note that this text replacement is generalized across every game in the GVGAI framework. As a result, the names of sprites (i.e. ``avatar (FlakAvatar)'' in Aliens) are displayed as they are written in VGDL. 

\begin{table}[tb]
\centering
\caption{Tutorial instructions for GVG-AI's Aliens}
\label{table:aliens-instructions}
\begin{tabular}{ll}
Controls: & As the avatar, use the arrow keys to turn and move.               \\ \cline{2-2} 
          & Press space to shoot the sam (missile).                           \\ \hline
Winning:  & If you press space, then avatar (FlakAvatar) will                        \\
          & shoot a sam (missile). \\ \cline{2-2} 
          & If alien and sam (missile) collide, then the alien                             \\
          & sprite will be destroyed.                               \\ \cline{2-2} 
          & If there are no more portalSlow (portal) sprites \\
          & or portalFast (portal) sprites or alienGreen (alien) \\
          & sprites or alienBlue (alien) sprites then you will win. \\ \hline
Losing:   & If avatar (FlakAvatar) and bomb (misisle) collide, \\
		  & then the avatar (FlakAvatar) sprite will be destroyed.    \\ \cline{2-2} 
		  & If avatar (FlakAvatar) and alien collide, then the \\
		  & avatar (FlakAvatar) sprite will be destroyed.    \\ \cline{2-2} 

          & If there are no more avatar (FlakAvatar sprites                               \\
          & then you will lose.                           \\ \hline
Points:   & If the alien and the sam (missile) collide, \\ 
		  &then you will gain 2 points.                                               \\ \cline{2-2} 
          & If the base and the sam (missile) collide,   \\
          & then you will gain 1 point.                   \\                                            
\end{tabular}
\end{table}

\subsection{Tutorial Videos}\label{sec:creating-video}
\begin{figure*}[tb]
  \begin{center}
      \includegraphics[width=2\columnwidth,height=0.35\textheight]{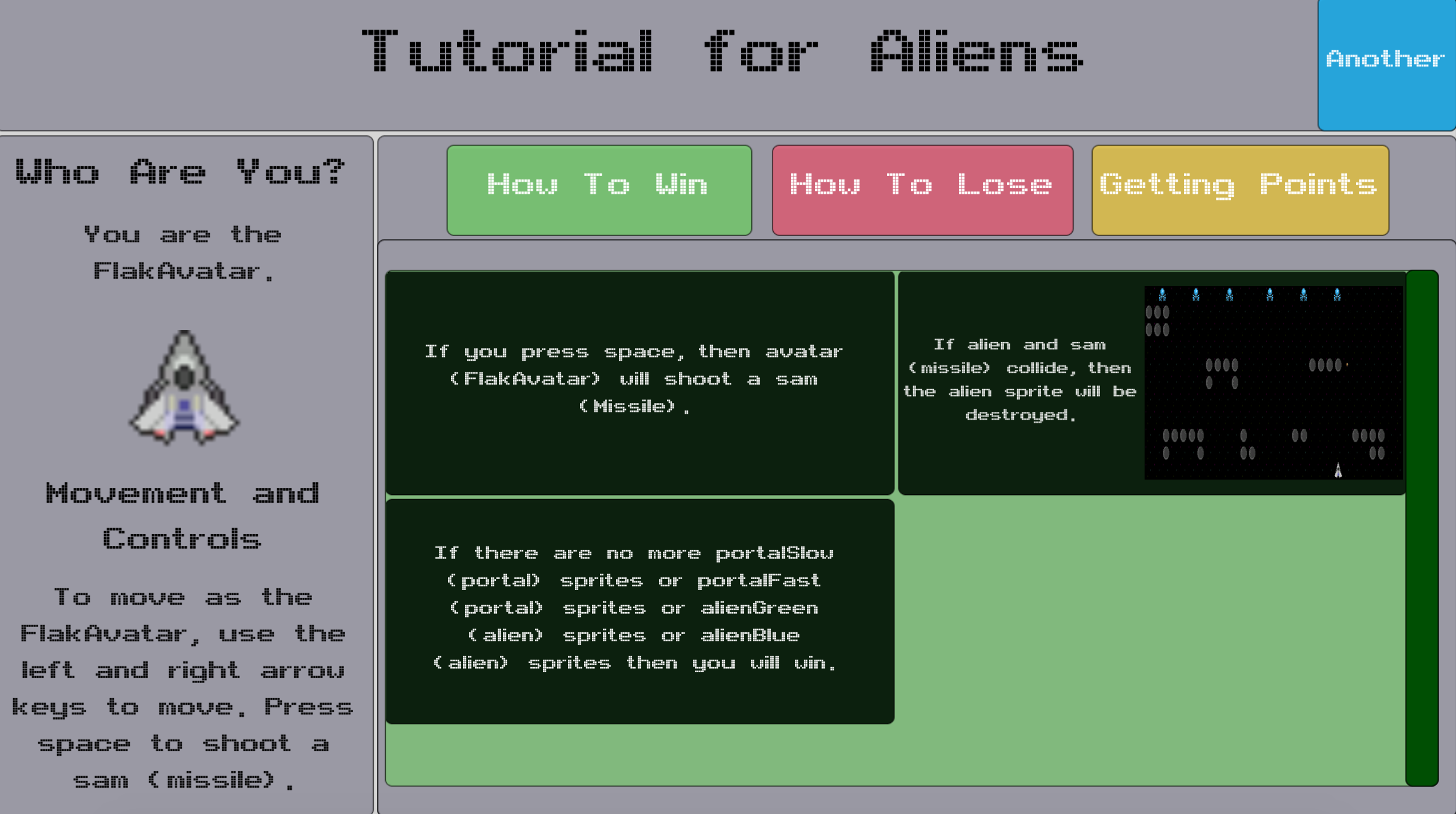}
  \end{center}
  \centering
\caption{A tutorial demonstrating mechanics for GVG-AI's Aliens, the buttons allow the user to switch from winning, losing, and point-gaining information}
\label{fig:tutorial-card-aliens}
\end{figure*}

Tutorial videos show animated examples of mechanics to the player
. To create them, the system, using an adapted version of \textit{Seekwhence}~\cite{Machado:2017:SRA:3102071.3102090}, takes frames captured by artificial agents playing the game and transform them into animations for the player to watch on a \emph{tutorial card}, which displays tutorial videos in an easy-to-read format. Figure~\ref{fig:tutorial-card-aliens} shows a section of such a card for GVG-AI's Aliens.

However, at the moment of this research there is yet to exist an agent that can beat every game in the GVG-AI framework. Therefore, in an attempt to make the generator as robust as possible, the system uses a cocktail of winning competitors from past GVG-AI competitions, all of which can be found on the GVG-AI website\footnote{www.gvgai.net}: adrienctx~\cite{perez20162014}, NovelTS, NovTea~\cite{geffner2015width}, Number27~\cite{perez2018general}, and YOLOBOT~\cite{joppen2017informed}. To capture the full extent of winning and losing conditions, our system also uses a one-step look-ahead agent and a doNothing agent. The former looks only one step ahead into the forward model, while the latter literally does nothing.

Each agent plays every level of a given game. 
Meanwhile, the system captures every frame and stores it in a database. After all agents have played, the system then queries the database for the exact frame where each mechanic in the critical paths and the points sections of the tutorial occurred. In the event that an agent was unable to beat a level, it might be missing frames for a winning critical path mechanic query, in which case the system will look at another playthrough. If an agent triggered this mechanic, the system requests 5 frames around the triggered mechanic to be placed in the tutorial video. Unused frames are deleted to save space. Mechanic frames found are bundled together with the mechanic they show and displayed on a \emph{tutorial card} in an easy to read format. Frames are looped through to create an animation demonstrating the mechanic. If no agent triggered a tutorial-referenced mechanic, the system does not display any frames. For mechanics that are merged together, the system displays a video for every mechanic that was merged.



\section{Evaluation}

To evaluate \emph{AtDELFI}, we used it to generate tutorials for 8 GVG-AI games: \emph{Aliens}, \emph{Butterflies}, \emph{Camelrace}, \emph{Jaws}, \emph{Plants}, \emph{RealPortals}, \emph{SurviveZombies}, and \emph{Zelda}. These games were selected from the four different groups identified by Bontrager et al.~\cite{bontrager2016matching}. This section is divided into \textit{readability of instructions and videos}, and \textit{analysis of metrics gathered during agent-play-through}.

\subsection{Readability}

The readability of generated tutorials was evaluated subjectively. For each tutorial, we read the generated instructions, observed the demonstrated mechanics, and compared it to the VGDL game description written by the GVG-AI game developers. Our evaluations are described below:

\begin{description}[style=unboxed,leftmargin=0.3cm]
\item[Aliens] Aliens is a clone of \emph{Space Invaders}. The player uses arrow keys to move left and right on the bottom of the screen, shooting down aliens and avoiding missiles they shoot back. The player score for each alien destroyed. If the player collides with an alien or an alien's missile, they lose the game. The Aliens generated tutorial was one of the most understandable among the 8 games. Information given in all four sections of the instructions adequately explained main characteristics of the game: destroy all aliens and do not get blown up. Agents collected frames that highlighted every mechanic, making it user-friendly.
\item[Butterflies] In Butterflies, players must collect all butterflies and keep them away from the flowers (as they eat the flowers). Each butterfly collected increases the score, and after collecting all butterflies, the player wins. If a butterfly eats a flower, it multiplies, and if all flowers are eaten the player loses. The game's challenge is balancing flower management to score high without losing. The Butterflies tutorial did an adequate job at explaining the objectives of the game. Instructions do not touch upon the strategy of letting butterflies multiply in order to maximize point gain, but it does explain how to win, how to lose, and that collecting butterflies is a way to increase score. In addition to winning, losing, and collecting butterflies for points, the videos show a sequence where a butterfly collides with a flower and multiplies, but it does not explicitly comment on this mechanic. It displays this information only to show that, by allowing all flowers to be destroyed, the player will lose the game.
\item[Camel Race] Camel Race is a racing game. To win, the player needs to be the first camel to touch the opposite side of the screen. There are various obstacles the player needs to avoid in order to accomplish this goal. The only way to gain any points is by winning the game. The generated tutorial explains the winning and losing mechanics. Agent videos show win and lose situations when the goal-line is on either side of the screen.
\item[Jaws] Jaws is a survival game where the player needs to evade/kill all sharks within 1000 ticks. The sharks are divided into two types, passive and aggressive. Passive sharks swim in a straight line and can be killed by the player's gun, while aggressive sharks swim towards the player and cannot be killed by bullets. When sharks die, they turn into jewels, which can be collected by the player for points. The instructions describe the winning and losing paths correctly. However, although they explain that by collecting jewels the score will increase, they fail to include that jewels are spawned after destroying a shark. This lack of detail could confuse a player, who does not immediately see where jewels come from. Only through watching a video can players see how jewels are spawned.
\item[Plants] Plants is a GVG-AI clone of \emph{Plants vs. Zombies} (PopCap Games 2009). The goal is to survive for 1000 ticks. Zombies spawn on the right side of the screen moving towards the left, and the player loses if a zombie reaches the left side of the screen. The player needs to build plants, which fire zombie-killing pea sprites. Generated instructions explain how to win and to lose, but come short when it comes to mechanics involving points: they only explain that when peas collide with zombies the score will increase. It does not explain that plants spawn peas or that the player can create plants to defend themselves. Additionally, the videos do include a sequence showing a pea spawning from a plant and hitting a zombie along side text ``If the zombie and the pea collide, then you will gain 1 point,'' but does not explicitly comment on the pea spawning mechanic.
\item[RealPortals] RealPortals is a GVG-AI clone of \emph{Portal} (Valve 2007). The player must reach the door, which sometimes is behind another locked door that needs a key. The player is restricted by water, which often lies between them and the door. To succeed, players need to pick up wands, which allow them to create \emph{portals} through which they can spontaneously travel across the map. There are two different types of wands, and each corresponds to a different portal gateway. The generated tutorial explain how to change portal types (by colliding with different wands), but does not explain how each wand is used to create portals, nor how the player can travel using the portals. The instructions fail to explain that the portals are necessary in order to win the game, however it does include that the player needs to collide with the goal to win. They also explain that using a portal gives the player points. None of the agents were able to beat RealPortals. They show an agent using portals to gain points, but always in compromising positions (e.g. the agent uses the portal to teleport into the water, proceeding to die immediately after). Because none of the agents won, the winning critical path information is incomplete, specifically the part about colliding with the goal to win the game, as seen in Figure \ref{fig:tutorial-card-realportals}.

\begin{figure*}[tb]
  \begin{center}
      \includegraphics[width=2\columnwidth,height=0.35\textheight]{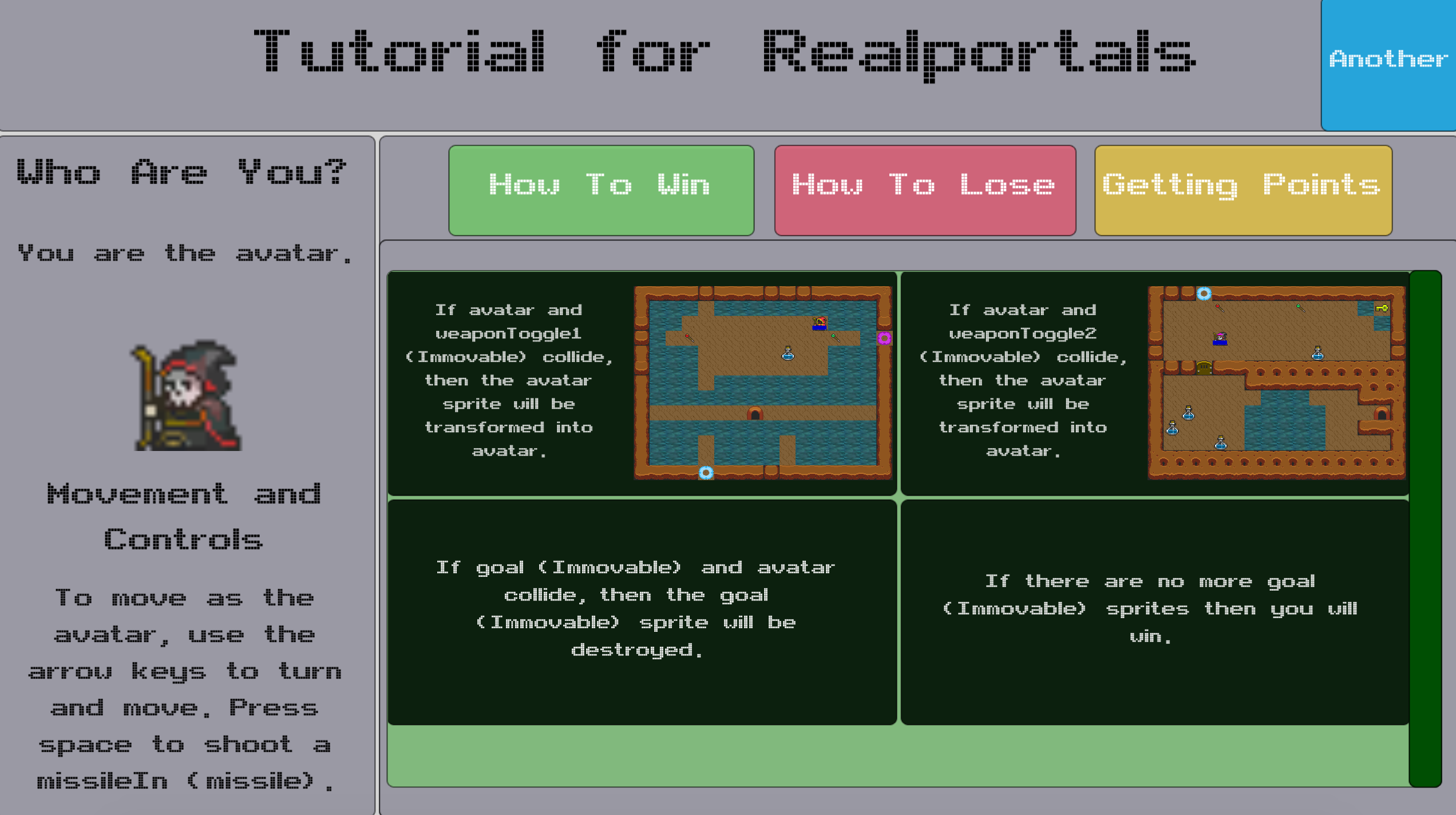}
  \end{center}
\caption{A tutorial demonstrating mechanics for GVG-AI's RealPortals, although there are agent videos for points and losing, the win section is incomplete due to the fact that none of the agents could beat the game.}
\label{fig:tutorial-card-realportals}
\end{figure*}
\item[SurviveZombies] SurviveZombies is a zombie-survival game, where the player needs to survive for 1000 ticks to win. The player can increase their score and get more hit points if they pick up honey, which is created when bee sprites collide with zombies. The player will lose hit points if they collide with any zombies, and if their hit points are depleted, they lose the game. The instructions describe surviving to win, but do not mention the hit point mechanic. This is due to this version of tutorial generation not being created with the understanding of hit point mechanics. It compensates for this by describing that collision with zombies is deadly: ``If avatar and hell collide, then the avatar sprite will be destroyed.'' The instructions also do not mention how honey regenerates hit points, only explaining that collecting honey awards points. However, the videos do show the hit point mechanic by displaying the moment hit points change when zombies are collided with or honey is collected.
\item[Zelda] Zelda is a simplified GVG-AI clone of the dungeon system in \emph{The Legend of Zelda} (Nintendo 1986). The goal is to pick up a key and unlock the door in the level. The player will encounter monsters, which can kill the player, causing them to lose. The player can swing a sword; if they hit a monster, the monster is destroyed, and the player gains points. Instructions explain the key and door mechanic used to win the game, as well as how the player can kill and be killed by monsters. The agent videos include sequences of every type of monster, killing and being killed by, the player, making this a user-friendly tutorial.
\end{description}

\subsection{Metrics}

\begin{figure}[b]
  \begin{center}
      \includegraphics[height=0.27\textwidth]{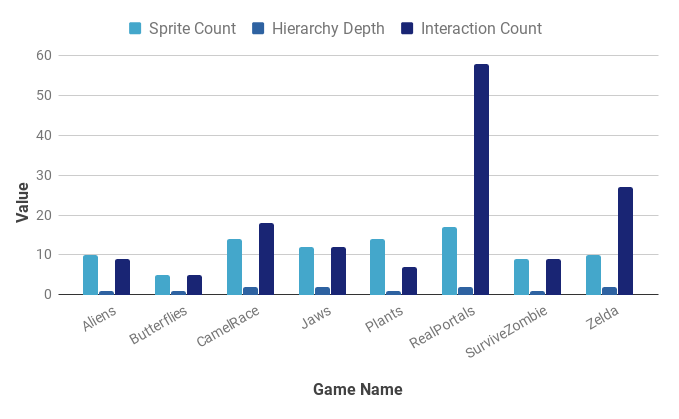}
  \end{center}
\caption{GVG-AI games statistics containing the amount of sprites, the maximum depth of parent hierarchy of sprites, and the amount of interactions in each game description.}
\label{fig:gamesAnalysis}
\end{figure}

Figure~\ref{fig:gamesAnalysis} shows statistics about different GVG-AI games. ``Sprite Count'' indicates the number of sprites that exist in a game's Sprite Set, ``Hierarchy Depth'' reflects the maximum depth of parent hierarchy of sprites, and ``Interaction Count'' is the number of interactions written in the Interaction Set. We see that the ``Sprite Count'' value is almost directly proportional to ``Interaction Count'' in all games except for Plants, RealPortals, and Zelda. Plants' ``Interaction Count'' is nearly half of its ``Sprite Count,'' whereas the ``Interaction Count'' is more than twice  the ``Sprite Count'' in Zelda. RealPortals is known in GVG-AI to be an extremely complicated game, and thus its ``Interaction Count'' is over three times its ``Sprite Count.''

\begin{figure}[tb]
  \begin{center}
      \includegraphics[height=0.27\textwidth]{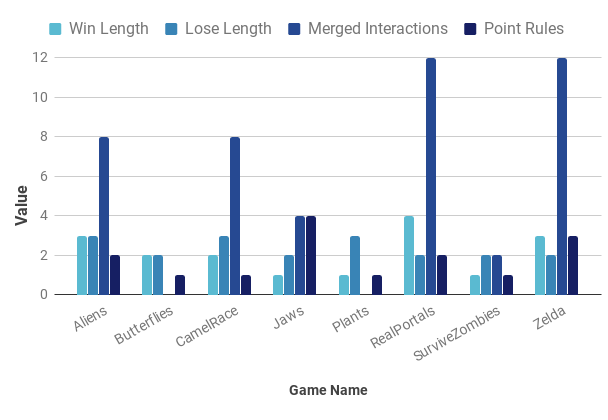}
  \end{center}
\caption{GVG-AI tutorial graphs statistics, showing the winning/losing path length, the amount of merged interactions in the mechanic graph and the amount of different ways to score points in each game.}
\label{fig:graphAnalysis}
\end{figure}

Figure \ref{fig:graphAnalysis} shows statistics about graphs generated from each game. ``Win Length'' tells the number of nodes on the winning critical path, ``Lose Length'' reflects the number of nodes on the losing critical path, ``Merged Interactions'' indicates how many interactions where merged into one, and ``Point Rules'' is the number of different ways the player can score. By looking at Figures \ref{fig:gamesAnalysis} and \ref{fig:graphAnalysis}, one might begin to conjecture why the tutorial for RealPortals was insufficient. We believe that the high ``Interaction Count'' (58 interactions) and the small number of ``Merged Interactions'' (12 interactions) leaves the algorithm with a rather large graph to traverse, compared to the mechanic graphs of other games.

\begin{figure}[tb]
  \begin{center}
      \includegraphics[height=0.27\textwidth]{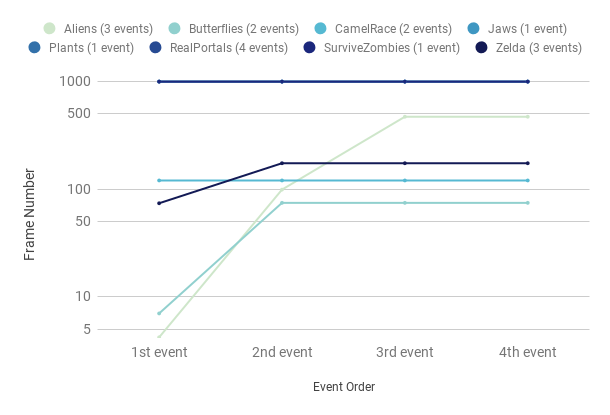}
  \end{center}
\caption{The average of the frame number where the winning critical path interactions happens for the first time.}
\label{fig:criticalPath}
\end{figure}

All agents track mechanics related to the winning critical path during play-through. Once all agents finish playing a specific game, we find, for all play-throughs resulting in victories, the first occurrence of each mechanic in the winning critical path. These are then averaged together by mechanic, by game. Figure \ref{fig:criticalPath} is a graph displaying where, on average, the winning critical path mechanics occur for the first time. The figure only shows a maximum of 4 events since this is the length of the longest winning critical path in any of these games. From the image, one can see that the frame number increases or remains the same with every next event. This demonstrates that if an agent won a game, it followed the order of the winning critical path. RealPortal is not displayed in the figure because none of the agents were able to reach a winning state. Jaws, Plants, and SurviveZombies are displayed as straight lines on top of each other because all of them have only one event in their critical path (``Win the game after 1000 ticks'').

\section{Conclusion}

Automatic tutorial generation is a non-trivial problem. Although our graph-based representation system works with simple arcade style games in the GVG-AI framework (such as Aliens and Zelda) when given an explicit rule representation, it breaks down when used on complex games like Plants or RealPortals. Tutorials for both of these games fail to adequately describe the game to the player. The Plants tutorial did not fully explain that the player needs to create plants, which fire pellets at the zombies. The RealPortals tutorial did not at all explain that the player can travel through the use of portals, or that using the portals to move around is required in every map of the game. This is because critical paths are built using just VGDL information in the current system, and the knowledge that one would \emph{have} to use portals to win would only be obtained by an actual play-through of the game. It is also of note to mention that none of the agents were able to win RealPortals, due to the large search space and delayed reward of winning. Because of this, its video tutorial missed frames for its winning section. To improve the system, we would need to either construct or use agents that are capable of winning the game in question.

It is obvious that there are some limitations to the simplistic graph system described in this paper. We believe that a successful tutorial generator needs to have a higher level of understanding than just simple mechanics, when it comes to more complex games. Our system used a longest-shortest path algorithm to find critical paths in the mechanic graph. However, we hypothesize that using an agent to track which mechanics it used during play would result in more effective tutorials. We did not implement this idea in this project due to time constraints, but by tracking the frequency count of each mechanic, we might be able to find a more accurate critical path containing well-used mechanics rather than a longest-shortest path, assuming that the mechanics the agent used more frequently are significant for winning.


We would like to explore tutorial generation without an explicit rule-based representation like VGDL. Rather, we want to discover mechanics by using agent play-throughs, and by using the discovered mechanics, construct a mechanic graph. Methods for automatically identifying maps, mechanics and other characteristics of games given only an executable version of the game show promise for this kind of project~\cite{osborn2017automatic}. As stated above, there are limitations to our system, and one priority is to improve it for GVG-AI games. We are interested in using this method on games outside of the GVG-AI framework, such as creating modular tutorials for various mechanics in \emph{Infinite Mario Bros} (Persson 2009) or building a treasure/scavenger hunt generator in \emph{Minecraft} (Mojang 2009).

\section*{Acknowledgment}
Michael Cerny Green acknowledges the financial support of the GAANN program. Ahmed Khalifa acknowledges the financial support from NSF grant (Award number 1717324 - ``RI: Small: General Intelligence through Algorithm Invention and Selection.''). Gabriella Barros acknowledges financial support from CAPES and Science Without Borders program, BEX 1372713-3, as well as an NYU Tandon School of Engineering Fellowship. Tiago Machado acknowledges the finacial support from CNPq - Conselho Nacional de Desenvolvimento Cient\'{i}fico e Tecnol\'{o}gico under the Science without Borders scholarship 202859/2015-0.

\bibliographystyle{ACM-Reference-Format}
\bibliography{bibliography}

\end{document}